\documentclass[pdflatex,sn-mathphys-num,iicol]{sn-jnl}

\usepackage{graphicx}%
\usepackage{multirow}%
\usepackage{amsmath,amssymb,amsfonts}%
\usepackage{amsthm}%
\usepackage{mathrsfs}%
\usepackage[title]{appendix}%
\usepackage{xcolor}%
\usepackage{textcomp}%
\usepackage{manyfoot}%
\usepackage{booktabs}%
\usepackage{algorithm}%
\usepackage{algorithmicx}%
\usepackage{algpseudocode}%
\usepackage{listings}%

\usepackage{placeins}

\newcommand{\best}[1]{\textbf{#1}}
\newcommand{\second}[1]{\underline{#1}}

\theoremstyle{thmstyleone}%
\theoremstyle{thmstyletwo}%

\theoremstyle{thmstylethree}%

\usepackage{ulem}

\begin{document}

\title[]{A Study of the Limits of Collaborative DCT-Based Image Denoising via Interpretable Neural Networks}


\author*[1]{\fnm{Cristian} \sur{Comellas}}\email{cristian.comellas@uib.es}

\author[1]{\fnm{Julia} \sur{Navarro}}\email{julia.navarro@uib.es}

\author[1]{\fnm{Antoni} \sur{Buades}}\email{toni.buades@uib.es}


\affil[1]{\orgdiv{Department of Mathematics and Computer Science}, \orgname{Universitat de les Illes Balears}, \orgaddress{\street{Cra. de Valldemossa, km 7.5}, \city{Palma}, \postcode{07122}, \state{Balearic Islands}, \country{Spain}}}


\abstract
{Image denoising remains a fundamental problem in image restoration, with applications in photography, biomedical, and scientific imaging. Modern deep neural networks achieve strong performance by learning powerful image priors, but often rely on large black-box models with limited interpretability. In contrast, DCT-based sliding-window and collaborative filtering methods such as BM3D offer clear algorithmic structure, but depend on handcrafted and non-differentiable operations. This work studies how far such structured collaborative filtering principles can be pushed when reformulated as trainable models. We introduce DeepBM3D, a compact fully differentiable architecture that combines non-local patch grouping, DCT-domain filtering, and multi-stage refinement within a BM3D-inspired pipeline. Lightweight convolutional feature extractors guide patch grouping, while filtering is performed through learned Wiener weights in the DCT domain. Experiments show that DeepBM3D improves over classical and hybrid baselines, remains competitive with FFDNet at low and moderate noise levels, and performs particularly well on repetitive textures.
}




\maketitle

\section{Introduction}
\label{sec:intro}

Image denoising is a long-standing problem in image processing and computer vision, fundamental to applications ranging from consumer photography to medical and scientific imaging. 
The task consists of recovering a clean image from a noisy observation, a challenge that lies at the intersection of statistical estimation, signal modeling, and machine learning. 
Classical methods based on sparsity and non-local redundancy, including transform-domain adaptive filtering approaches~\cite{Yaroslavsky1998}, Wiener filtering~\cite{lim1990wiener}, and collaborative filtering methods such as BM3D~\cite{dabov2007bm3d}, have remained reference points for decades thanks to their interpretability, efficiency, and strong performance despite relying on handcrafted priors. 
These algorithms exploit structured operations such as transform-domain filtering, often relying on transforms like the DCT, together with patch grouping to leverage sparsity and self-similarity within natural images, providing a clear understanding of each processing stage.

Deep learning has fundamentally changed the landscape of image restoration. 
Modern convolutional and transformer-based models~\cite{zhang2017dncnn,zhang2018ffdnet,zamir2022restormer} learn powerful image priors directly from data, achieving state-of-the-art performance across diverse noise conditions and datasets. 
However, these gains often come at the expense of interpretability and computational efficiency. 
Typical networks contain millions of parameters and operate as black boxes, making their internal representations difficult to analyze or control. 
In contrast, classical denoising pipelines offer a clear algorithmic structure and modularity, which explicitly separates the grouping, filtering, and aggregation stages.

This growing contrast has motivated renewed interest in models that bridge the gap between data-driven flexibility and the principled structure of traditional algorithms. Recent model-based and hybrid approaches~\cite{plotz2018n3net,wang2018nonlocalnet,ye2024ilrnet} have shown that it is possible to embed optimization principles and classical operations into trainable networks, combining interpretability with learning. In particular, Herbreteau et al.~\cite{herbreteau2022dct2net} proposed DCT2Net, demonstrating that classical DCT-based denoising pipelines can benefit from learnable thresholding strategies. 

Building upon these ideas, we propose DeepBM3D, a compact and fully differentiable architecture for collaborative transform-domain image denoising. DeepBM3D replaces handcrafted components commonly used in classical collaborative filtering pipelines with learnable and fully differentiable modules. Our model adopts a two-stage design and introduces lightweight convolutional modules to guide patch grouping and estimate transform-domain filtering weights in a trainable manner. This formulation enables gradient-based optimization across all stages while preserving the structural interpretability of collaborative filtering approaches. 

To preserve the underlying structure of classical collaborative filtering pipelines, we use a fixed DCT as transform and PSAL \cite{cherel2024psal}, which performs differentiable patch matching. We explore two configurations: a lightweight version that already outperforms several classical and hybrid methods, and a heavier variant that approaches or exceeds FFDNet~\cite{zhang2018ffdnet}, a representative CNN-based denoiser without explicit collaborative filtering constraints, in selected low and moderate noise settings.

Our results demonstrate that classical collaborative filtering principles, including non-local patch grouping and transform-domain Wiener filtering, can be effectively integrated within a fully differentiable and end-to-end trainable framework. The proposed architecture combines interpretability, modularity, and competitive denoising performance within a compact trainable model.

The remainder of this paper is organized as follows. 
Section~\ref{sec:sota} reviews related classical, deep, and hybrid denoising approaches. 
Section~\ref{sec:bm3d} briefly summarizes the key principles of collaborative filtering as exemplified by BM3D~\cite{dabov2007bm3d}.
Section~\ref{sec:deepbm3d} presents our proposed DeepBM3D architecture. 
Experimental setup and results are described in Sections~\ref{sec:expsetup} and~\ref{sec:results}, respectively. 
Finally, Section~\ref{sec:conclusion} concludes the paper and discusses future research directions.


\section{Related Work}
\label{sec:sota}

\noindent
This section reviews the main methodological trends shaping the evolution from classical signal processing and non-local filtering, through modern deep architectures, to hybrid frameworks that integrate optimization and learning. 
Together, these developments motivate the development of structured, learnable denoising architectures that combine classical priors with deep learning.

\subsection{Classical Image Denoising Methods}
\label{sec:classical}

\noindent
Early image denoising methods relied on signal processing and statistical models that enforced sparsity, smoothness, or self-similarity. 
Sliding DCT thresholding ~\cite{Yaroslavsky1998} and Wavelet thresholding~\cite{donoho1995wavelet} exploited sparsity of coefficients to suppress noise while preserving edges, whereas dictionary-learning methods such as K-SVD~\cite{aharon2006ksvd} represented patches as sparse combinations of learned atoms, later extended through online learning for scalability~\cite{mairal2009onlinesparse}. 
The Non-Local Means (NLM) algorithm~\cite{buades2005nlm} introduced the idea of self-similarity by averaging non-local patches with similarity-based weights, a principle refined by the Non-Local Bayes model~\cite{lebrun2013nlbayesdenoising}, which framed patch grouping and filtering in a Bayesian setting. 
Low-rank modeling further generalized these ideas by assuming that groups of similar patches form low-rank matrices. 
Weighted Nuclear Norm Minimization (WNNM)~\cite{gu2014wnmm,gu2014wnmm_denoising} formalized this as a convex optimization problem with adaptive weighting of singular values, effectively linking patch grouping and sparse representation.

Other frameworks approached denoising from a probabilistic viewpoint. 
The Expected Patch Log Likelihood (EPLL) model~\cite{zoran2011probmodel} used Gaussian mixture models to describe natural image patches, achieving strong generalization across noise levels, while the Wiener filter~\cite{lim1990wiener} provided a theoretical foundation for optimal linear denoising under Gaussian assumptions. 
Among these approaches, collaborative filtering methods based on non-local patch grouping and transform-domain sparsity have proven particularly effective. 
A prominent example is the Block-Matching and 3D Filtering (BM3D) algorithm~\cite{dabov2007bm3d,lebrun2012bm3d}, which combines non-local grouping, transform-domain filtering, and patch aggregation in a unified multi-stage process.
BM3D’s success lies in exploiting both non-local redundancy and transform-domain sparsity, yielding highly competitive results even compared to modern learning-based models. 
Extensions such as BM3D-PCA~\cite{dabov2009bm3dPCAadaptive} and BM4D~\cite{maggioni2023bm4d} adapted the framework to data-dependent bases and volumetric signals. 
However, the reliance on handcrafted transforms and hard block matching limits adaptability, motivating the transition toward trainable and fully differentiable alternatives.

\subsection{Learning-Based Denoisers: From CNNs to Transformers}
\label{sec:dl}

\noindent

Deep learning methods replaced image priors with learned representations, enabling powerful data-driven denoising. 
Early neural approaches used multilayer perceptrons~\cite{burger2012mlpdenoiser}, but convolutional neural networks (CNNs) soon became dominant due to their parameter sharing and spatial locality. 
DnCNN~\cite{zhang2017dncnn} pioneered residual learning by predicting the noise component instead of the clean image, and FFDNet~\cite{zhang2018ffdnet} incorporated a noise-level map to handle spatially variant noise with a single model. 
MemNet~\cite{tai2017memnet} introduced memory blocks to retain contextual information across layers, and subsequent works generalized these ideas to diverse restoration tasks.

Self-supervised and unsupervised paradigms addressed the scarcity of clean training data. 
Noise2Noise~\cite{lehtinen2018noise2noise} demonstrated that denoisers can be trained using only pairs of noisy images, while Noise2Void~\cite{krull2019noise2void} and Noise2Self~\cite{batson2019noise2self} extended this principle to single-image denoising through pixel masking strategies. 
These methods significantly broadened the applicability of deep denoising in real-world imaging scenarios.

The encoder–decoder architecture introduced by U-Net~\cite{ronneberger2015unet} became the foundation for most CNN-based denoisers. 
Variants such as RIDNet~\cite{anwar2019ridnet}, DDUNet~\cite{jia2021ddunet}, and SADNet~\cite{chang2020sadnet} integrated attention mechanisms, dense connectivity, and spatial adaptivity to enhance feature representation. 
Later architectures like MPRNet~\cite{zamir2021mprnet}, HINet~\cite{chen2022hinet}, and NAFNet~\cite{chen2022nafnet} employed multi-stage refinement and improved normalization for higher efficiency and stability. 
These advances consolidated CNNs as the standard for high-quality image restoration. 
Beyond task-specific CNNs, general-purpose denoisers such as DRUNet~\cite{zhang2021drunet} have been proposed as flexible modules that can serve both as standalone restoration networks and as plug-and-play priors for a wide range of inverse problems. 
This design demonstrates how a single deep architecture can act as a universal regularizer, bridging traditional optimization frameworks and data-driven learning.

Transformers extended this progress by modeling long-range dependencies. 
SwinIR~\cite{liang2021swinir} builds upon the Swin Transformer~\cite{liu2021swintransformer} architecture, adapting its shifted-window self-attention mechanism for efficient and scalable image restoration. 
Restormer~\cite{zamir2022restormer} and Uformer~\cite{wang2022uformer} further extend this paradigm by combining hierarchical attention with U-shaped architectures, enabling global context modeling while preserving spatial precision.

FFTFormer~\cite{zhu2025fftformer} further bridged spatial and frequency domains, and DnT~\cite{li2022dnt} proposed an unsupervised transformer-based denoiser. 
Together, these models demonstrate a shift from explicit priors to hierarchical learned representations. 
More recently, diffusion-based generative models have also been adapted for image restoration. 
SR3~\cite{saharia2023diffusion} formulates super-resolution as an iterative denoising process, showing that diffusion mechanisms can progressively refine images toward high-fidelity reconstructions. 
Although primarily developed for generative tasks, these models highlight the close conceptual link between denoising and iterative refinement.
Efficiency-oriented designs, such as EfficientNet~\cite{tan2019efficientnet}, also influenced modern lightweight denoisers for mobile and real-time applications.

\subsection{Model-Based Deep Learning and Hybrid Paradigms}
\label{sec:hybrid}

\noindent
Model-based deep learning seeks to combine interpretability with learning capacity by embedding optimization principles into neural architectures. 
Algorithm unrolling~\cite{gregor2010LearningFA} first showed that sparse coding inference can be approximated through a fixed-depth network, inspiring frameworks such as ADMM-Net~\cite{yang2017admmnet} and ADMM-CSNet~\cite{yang2020admmcsnet}, which unroll iterative solvers for compressive sensing and MRI. 
Similarly, TNRD~\cite{chen2017tnrd} learned parameters of PDE-based diffusion processes, linking deep networks and variational models. 
Plug-and-Play (PnP)~\cite{Venkatakrishnan2013pnp} and Regularization by Denoising (RED)~\cite{romano2017red} further connected classical optimization with deep priors, treating denoisers as implicit regularizers. 
Related approaches learned explicit proximal operators~\cite{meinhardt2017proxops} or exploited network structure as an implicit prior, as in Deep Image Prior~\cite{lempitsky2018dip}.

Hybrid architectures integrate classical operations directly within neural frameworks. 
BM3D-Net~\cite{yang2018bm3dnet} and DCT2Net~\cite{herbreteau2022dct2net} emulate transform-domain filtering, while BMCNN~\cite{ahn2018bmcnn} incorporates block-matching as a learnable stage. 
Differentiable non-local modules such as N3Net~\cite{plotz2018n3net}, Non-Local Networks~\cite{wang2018nonlocalnet}, and Dual Attention~\cite{fu2019dualattention} generalize patch similarity into attention mechanisms, later extended by PSAL~\cite{cherel2024psal}, which performs differentiable approximate nearest-neighbor matching via stochastic attention.
Low-rank priors have also been integrated into deep unrolled architectures, as in HLR-DUR~\cite{shao2024hlr} and ILRNet~\cite{ye2024ilrnet}, bridging low-rank modeling and deep optimization. 
Recent reviews~\cite{su2022dlsurveyir,elad2023dlsurvey} summarize this convergence between classical and deep paradigms, emphasizing that denoisers can serve as flexible priors for inverse problems.

\subsection{Learning Collaborative Transform-Domain Denoising}
\label{sec:related_method}

\noindent
Collaborative filtering methods based on non-local patch grouping and transform-domain sparsity have long been central to classical image denoising, with BM3D~\cite{dabov2007bm3d,lebrun2012bm3d} representing one of the most influential examples. 
These approaches inspired several attempts to integrate their principles within learnable frameworks. 
Methods such as BM3D-Net~\cite{yang2018bm3dnet} and BMCNN~\cite{ahn2018bmcnn} introduce neural components into collaborative filtering pipelines, while still relying on fixed or non-differentiable stages. 
Other works focus on learning individual components of these pipelines: DCT2Net~\cite{herbreteau2022dct2net} learns interpretable transform-domain filters, N3Net~\cite{plotz2018n3net} approximates patch matching through attention mechanisms, and PSAL~\cite{cherel2024psal} performs differentiable patch grouping via stochastic attention.

These developments highlight the potential of combining non-local similarity, transform-domain filtering, and deep learning within unified architectures. 
Building on these ideas, the proposed DeepBM3D integrates non-local patch grouping and collaborative transform-domain filtering into a fully differentiable and trainable framework, enabling end-to-end learning while preserving the structural principles of classical collaborative denoising methods.


\section{Key Principles of BM3D}
\label{sec:bm3d}

\noindent
Block-Matching and 3D Filtering (BM3D)~\cite{dabov2007bm3d,lebrun2012bm3d} is a classical image denoising algorithm that exploits non-local self-similarity through collaborative filtering in a transform domain. 
In this section we briefly summarize the key principles of BM3D that are relevant for understanding the proposed architecture.

Let $Y = X + \eta$ denote the observed noisy image, where $X$ is the clean image and $\eta \sim \mathcal{N}(0,\sigma^2)$ is additive white Gaussian noise. 
BM3D operates by grouping similar patches, applying collaborative filtering in a transform domain, and aggregating the resulting estimates.

\bigskip
\noindent\textbf{Patch Grouping}
For each reference patch extracted from the noisy image, BM3D searches within a spatial window to identify a set of similar patches according to an $\ell_2$ distance. 
These patches are stacked together to form a 3D group, exploiting the observation that similar structures often recur across an image.

\bigskip 
\noindent\textbf{Collaborative Transform-Domain Filtering}
Each group is processed using a separable 3D transform that combines a 2D transform applied to each patch and a 1D transform across the group dimension. 
Noise is suppressed by shrinking transform coefficients, either through hard thresholding or Wiener filtering, before applying the inverse transform to reconstruct denoised patches.

\bigskip 
\noindent\textbf{Aggregation}
The filtered patches are returned to their original image locations. 
Because patches overlap, multiple estimates contribute to each pixel. 
A weighted averaging strategy aggregates these contributions, producing the final restored image.

BM3D performs this procedure in two successive stages: a first stage based on hard-thresholding to obtain a basic estimate, followed by a second stage that refines the result using Wiener filtering guided by the basic estimate. 
These principles of non-local patch grouping, transform-domain collaborative filtering, and aggregation form the conceptual foundation for the architecture proposed in this work.


\section{DeepBM3D Architecture}
\label{sec:deepbm3d}

\noindent We introduce \textbf{DeepBM3D}, a lightweight end-to-end trainable architecture for collaborative transform-domain image denoising. 
The proposed model integrates non-local patch grouping, transform-domain collaborative filtering, and aggregation within a unified differentiable framework. 
Its design is guided by the structural principles of classical collaborative filtering methods, including BM3D,  with trainable modules that enable data-driven optimization.

\begin{figure*}[!t]
    \centering
    \includegraphics[width=\linewidth]{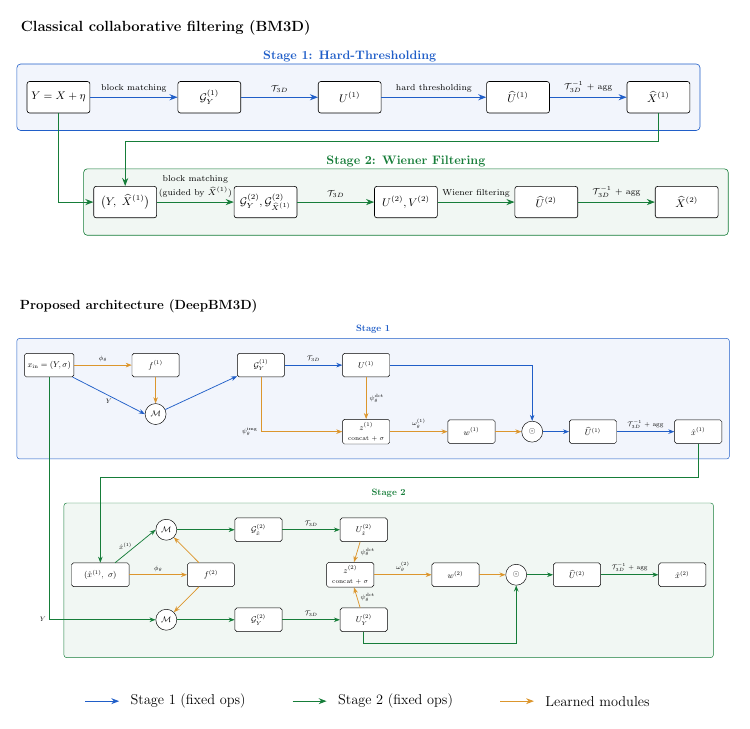}
    \caption{Comparison of BM3D (top) and DeepBM3D (bottom): both follow a two-stage design, with fixed operators shown in blue/green and learned modules in orange. The operator $\mathcal{M}$ represents block matching using PSAL~\cite{cherel2024psal}.}
    \label{fig:bm3d_deepbm3d}
    
\end{figure*}

DeepBM3D adopts a two-stage architecture that follows the progressive refinement strategy commonly used in classical denoising pipelines. 
In this design, three key operations are implemented as differentiable modules: patch grouping is achieved through an attention-based mechanism (PSAL~\cite{cherel2024psal}), collaborative filtering is performed via learned Wiener weights in the DCT domain, and the 3D transform is implemented as a linear layer. 
In addition, compact CNN-based feature extractors are employed to guide the grouping process and enhance representational power.
A visual comparison between BM3D and DeepBM3D is shown in Figure~\ref{fig:bm3d_deepbm3d}. 
Overall, this yields a modular and interpretable architecture that combines the structural principles of collaborative filtering with the flexibility of deep learning.

\subsection{Mathematical formulation}

\noindent Let $Y \in \mathbb{R}^{1 \times H \times W}$ be a noisy grayscale input image, with $Y = X + \eta$ where $\eta \sim \mathcal{N}(0, \sigma^2)$. A noise level map $\sigma \in \mathbb{R}^{1 \times H \times W}$ is assumed known and concatenated channel-wise with the input:
\[
x_{\text{in}} = \texttt{cat}(Y, \sigma) \in \mathbb{R}^{2 \times H \times W}.
\]

The architecture is structured around the following modules: 
(i) a fixed, differentiable block-matching operator $\mathcal{M}$ implemented with PSAL, 
(ii) three feature extractors: $\phi_\theta$ for image-level features, $\psi_\theta^{\text{img}}$ for patch-level spatial features, and $\psi_\theta^{\text{dct}}$ for patch-level transform features, 
(iii) two Wiener weight estimation networks $\omega_\theta^{(1)}$ and $\omega_\theta^{(2)}$, and 
(iv) fixed 3D DCT/IDCT operators implemented as separable 2D and 1D transforms. 

\paragraph{Stage 1} 
The first step, as in BM3D, is patch grouping. To guide this process, feature maps are extracted from the noisy input and the given noise map:
\[
f^{(1)} = \phi_\theta(x_{\text{in}}).
\]
For each reference location $i$, the block-matching module $\mathcal{M}$ receives two inputs:  
(i) a feature tensor $f^{(1)}$ that provides a learned representation for measuring similarity, and  
(ii) the image $Y$, from which the actual patches are extracted.  
The operator selects the $C$ most similar patches to the reference one according to distances in feature space, 
and then gathers the corresponding patches from $Y$, producing
\[
\mathcal{G}_{Y,i}^{(1)} = \mathcal{M}(f^{(1)}_i,\, Y), \quad 
\mathcal{G}_{Y,i}^{(1)} \in \mathbb{R}^{N \times k \times k}, \quad N = C + 1.
\]
This follows the same principle of non-local patch grouping used in classical collaborative filtering methods (where $N$ denotes the number of patches per group), but here the selection of matching coordinates is guided by PSAL, 
which performs differentiable, attention-based matching in the feature domain.  
Although PSAL itself has no trainable parameters, it enables gradient propagation through patch selection, a key advantage over traditional hard matching.

Each group is transformed with a separable 3D DCT:
\[
U_{Y,i}^{(1)} = \mathcal{T}_{3D}(\mathcal{G}_{Y,i}^{(1)}).
\]

Filtering weights are predicted and applied:
\[
w_i^{(1)} = \omega_\theta^{(1)}(z_i), \quad \widehat{U}_{Y,i}^{(1)} = w_i^{(1)} \odot U_{Y,i}^{(1)}, 
\]
where $\odot$ represents the Hadamard product, and the feature vector $z_i$ is generated as follows:
\begin{align*}
    z_{\text{dct}} &= \psi_\theta^{\text{dct}}(U_{Y,i}^{(1)}), \\
    z_{\text{img}} &= \psi_\theta^{\text{img}}(\mathcal{G}_{Y,i}^{(1)}), \\
    z_i &= \texttt{cat}(z_{\text{dct}}, z_{\text{img}}, \sigma).
\end{align*}

After the inverse transform, the filtered patches are aggregated at their reference locations to form the first-stage output:
\[
\hat{x}^{(1)} = \text{Aggregate}(\mathcal{T}_{3D}^{-1}(\widehat{U}_{Y,i}^{(1)})).
\]
Although $\hat{x}^{(1)} \in \mathbb{R}^{1 \times H \times W}$ has the shape of an image, it is not constrained to visually resemble $X$. No supervision is applied at this stage, and $\hat{x}^{(1)}$ should be interpreted as a latent intermediate representation optimized to benefit the second stage. This output plays the same functional role as $\widehat{X}^{(1)}$ in classical BM3D, serving as a guidance signal for the second stage.

\paragraph{Stage 2}
The second stage mirrors this process but leverages the intermediate estimate for more reliable grouping and filtering. Features are first extracted from the concatenation of the basic estimate and the noise map:
\[
f^{(2)} = \phi_\theta(\texttt{cat}(\hat{x}^{(1)}, \sigma)).
\]

PSAL is then reused to group patches from both the noisy input $Y$ and the intermediate estimate $\hat{x}^{(1)}$:
\begin{align*}
    \mathcal{G}_{Y,i}^{(2)} &= \mathcal{M}(f^{(2)}_i, Y), \\
    \mathcal{G}_{\hat{x},i}^{(2)} &= \mathcal{M}(f^{(2)}_i, \hat{x}^{(1)}),
\end{align*}

with $\mathcal{G}_{Y,i}^{(2)}, \mathcal{G}_{\hat{x},i}^{(2)} \in \mathbb{R}^{N \times k \times k}$.

Each group is transformed with the separable 3D DCT:
\[
U_{Y,i}^{(2)} = \mathcal{T}_{3D}(\mathcal{G}_{Y,i}^{(2)}), \qquad
U_{\hat{x},i}^{(2)} = \mathcal{T}_{3D}(\mathcal{G}_{\hat{x},i}^{(2)}).
\]

Wiener weights are predicted and applied elementwise:
\[
w_i^{(2)} = \omega_\theta^{(2)}(z_i^{(2)}), \qquad
\widehat{U}_{Y,i}^{(2)} = w_i^{(2)} \odot U_{Y,i}^{(2)},
\]

where the feature vector $z_i^{(2)}$ is encoded from the transform domain and concatenated with the noise map:
\begin{align*}
    h_{Y,i} = \psi_\theta^{\text{dct}}(U_{Y,i}^{(2)}), \\
    h_{\hat{x},i} = \psi_\theta^{\text{dct}}(U_{\hat{x},i}^{(2)}), \\
    z_i^{(2)} = \texttt{cat}(h_{Y,i}, h_{\hat{x},i}, \sigma).
\end{align*}

Finally, inverse transforms and aggregation at reference locations yield the denoised estimate:
\[
\hat{x}^{(2)} = \text{Aggregate}\!\left(\mathcal{T}_{3D}^{-1}\!\left(\widehat{U}_{Y,i}^{(2)}\right)\right),
\]

where the final estimate $\hat{x}^{(2)}$ is therefore analogous to $\widehat{X}^{(2)}$ in BM3D.

\subsection{Implementation details}

\noindent The learned modules in DeepBM3D are designed to be lightweight while retaining interpretability. 
All convolutional layers employ $3 \times 3$ kernels with \texttt{reflect} padding to mitigate boundary artifacts, followed by ReLU activations unless otherwise noted.

\begin{figure*}
\centering
 
\includegraphics[width=0.7\linewidth]{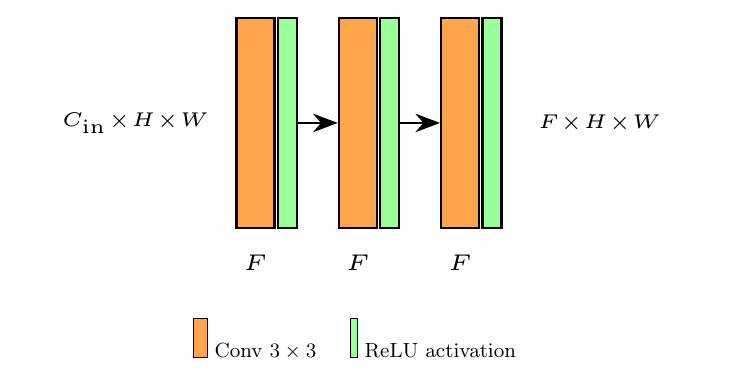}

\caption{Feature-extraction block used across DeepBM3D ($\phi_\theta$, $\psi_\theta^{\text{img}}$, and $\psi_\theta^{\text{dct}}$). The module is three
consecutive \emph{Conv $3{\times}3$ + ReLU} layers with \texttt{reflect}
padding, thus preserving the spatial size $(H{\times}W)$.  For the
image-level extractor $\phi_\theta$, the channel sizes are
$C_{\text{in}}{=}2$ (noisy image and noise map) and
$F{=}32$ (\emph{light}) / $F{=}64$ (\emph{heavy}).  For the patch-level
extractors $\psi_\theta^{\mathrm{img}}$ and $\psi_\theta^{\mathrm{dct}}$,
$C_{\text{in}}{=}N$ candidates with $N {=} C {+} 1$, and
$F {=} Nm$ with $m{=}4$ (\emph{light}) / $m{=}8$ (\emph{heavy}).}
\label{fig:feature-extractor}
\end{figure*}

\bigskip
\noindent\textbf{Feature extractors} 
DeepBM3D relies on three small CNNs that share a common architecture (see Figure~\ref{fig:feature-extractor}) but serve distinct roles: 
(i)~the image-level extractor $\phi_\theta$, which operates on the concatenation of the noisy image and the noise map; 
(ii)~the spatial-domain extractor $\psi_\theta^{\text{img}}$, applied to grouped patches in the pixel domain; 
and (iii)~the transform-domain extractor $\psi_\theta^{\text{dct}}$, applied to grouped patches in the DCT domain. 
Each extractor follows the structure Conv $3\times3 \rightarrow$ Conv $3\times3 \rightarrow$ Conv $3\times3$, with \texttt{reflect} padding and ReLU activations after each convolution. 
The output dimensionality of $\phi_\theta$ is controlled by $n_\text{features}$, while $\psi_\theta^{\text{img}}$ and $\psi_\theta^{\text{dct}}$ produce $N \cdot m$ features, where $N$ is the number of grouped patches and $m$ a capacity multiplier. 
Two configurations are considered:  
\begin{itemize}
    \item \textit{light:} $n_\text{features}=32$, $m=4$.  
    \item \textit{heavy:} $n_\text{features}=64$, $m=8$.  
\end{itemize}

\bigskip
\noindent\textbf{Block matching} 
Patch grouping is implemented with the differentiable PSAL operator~\cite{cherel2024psal}. 
Patch similarity is computed on learned features, while the grouped patches are drawn from either the noisy image $Y$ or the first-stage estimate $\hat{x}^{(1)}$. 
This ensures that reconstruction is always grounded in the input, while gradients propagate through the selection mechanism.
The number of candidates is $C$, plus the reference patch, yielding $N=C+1$ patches per group. 
In practice, we use patch size $k=5$, and $N=16$ candidates.

\bigskip
\noindent\textbf{3D transforms} 
Groups are processed with separable 2D+1D DCT/IDCT operators implemented as fixed, non-trainable linear layers. 
This retains the transform-domain formulation commonly used in collaborative filtering methods, while ensuring differentiability and a low parameter count. 

\begin{figure*}
\centering

\includegraphics[width=\linewidth]{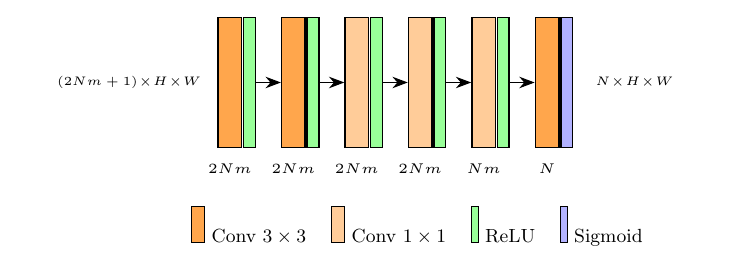}

\caption{Wiener weights estimation network ($\omega_\theta^{(1)}$ and $\omega_\theta^{(2)}$). The model combines \emph{Conv $3\times3$}, \emph{Conv $1\times1$}, and nonlinear activations to map features of size $(2Nm{+}1)\!\times\!H\times\!W$ to Wiener weights of size $N\!\times\!H\!\times\!W$. All convolutions use \texttt{reflect} padding. Here, $N$ denotes the number of grouped patches and $m$ the internal capacity multiplier. In the \emph{light} configuration, $m{=}4$; in the \emph{heavy} configuration, $m{=}8$.}

\label{fig:wiener-net}
\end{figure*}

\bigskip
\noindent\textbf{Wiener weights} 
Collaborative filtering is parameterized by two identical CNNs, $\omega_\theta^{(1)}$ and $\omega_\theta^{(2)}$, which estimate elementwise weights for Stage~1 and Stage~2, respectively. 
Each network follows the structure 
Conv $3\times3 \rightarrow$ Conv $3\times3 \rightarrow$ Conv $1\times1 \rightarrow$ Conv $1\times1 \rightarrow$ Conv $1\times1 \rightarrow$ Conv $3\times3 \rightarrow$ sigmoid, 
with ReLU activations between convolutional layers and \texttt{reflect} padding to reduce boundary artifacts. The full architecture is shown in Figure~\ref{fig:wiener-net}.
Outputs are bounded in $[0,1]$ by the final sigmoid activation but are not normalized across candidates, granting the model additional flexibility. 
Both \textit{DeepBM3D-light} and \textit{DeepBM3D-heavy} use this same structure, differing only in their respective multipliers ($m=4$ and $m=8$).

\bigskip
\noindent\textbf{Aggregation} 
We adopt reference patch replacement. Each denoised reference patch substitutes its noisy counterpart, and overlapping contributions are averaged. 
Alternative schemes, such as weighted overlapping or central-pixel aggregation, were tested but did not improve results. 

\bigskip
\noindent\textbf{Inference}
To enhance robustness and reduce boundary artifacts, the model processes the image in a sliding-window fashion over overlapping patches. For a patch size of $64 \times 64$, we use a stride of 4 during evaluation, although good results are also obtained with stride 32. The outputs of overlapping regions are merged using simple averaging. In practice, the formulation above applies equally whether $Y$ denotes a full image or an individual patch.


\section{Experimental setup}
\label{sec:expsetup}

\noindent We describe the experimental protocol used to evaluate DeepBM3D, including the datasets and noise configurations, training procedure, and baseline methods used for comparison.

\subsection{Datasets and Noise Settings}
 
\noindent To train and validate our models, we construct a grayscale dataset by combining images from DIV2K~\cite{timofte2017div2k}, Flickr2K~\cite{lim2017enhanced}, and the Waterloo Exploration Database~\cite{ma2017waterloo}. All images are converted to grayscale, resulting in a fixed split of 7{,}969 training images and 292 validation images, which is used consistently across all experiments. During training, random crops of size \(64 \times 64\) are sampled from these images at each epoch and used as input patches. In addition, data augmentation is applied in the form of random horizontal and vertical flips, as well as random $90^\circ$ rotations, to improve generalization. This dataset is used exclusively for training and validation.

For testing, we follow common practice and evaluate on four standard benchmarks: BSD100~\cite{bsd100}, Kodak~\cite{kodak}, McMaster (McM)~\cite{mcm}, and Set14~\cite{set14}. These datasets provide a diverse range of content and are widely adopted for denoising evaluation.

Noise is synthetically applied as additive white Gaussian noise (AWGN). During flexible training phases (see Section~\ref{sec:train_details}), noise levels \(\sigma\) are sampled uniformly from the interval \([0, 50]\). For evaluation, we report results at fixed noise levels \(\sigma \in \{5, 15, 25, 35\}\), which are standard in the denoising literature. All noise realizations are generated using a fixed random seed to ensure reproducibility, so that test conditions are identical across runs.

\subsection{Training Details}
\label{sec:train_details}

\noindent We adopt a three-phase training scheme to optimize both generalization and performance across noise levels. 
All models are trained end-to-end using an $\ell_1$ loss on the final output $\hat{x}^{(2)}$. 
Optimization is performed with Adam ($\text{lr}=10^{-4}, \beta_1=0.9, \beta_2=0.999$), using an effective batch size of 16. 
Training is conducted on a single NVIDIA RTX A6000 GPU (48 GB).

During training, input patches are sampled dynamically from the training images. 
At each epoch, random $64\times64$ crops are drawn from randomly selected images, so that only a subset of the dataset is seen per epoch. 
This stochastic sampling strategy effectively exposes the model to a large diversity of patches across epochs while keeping each epoch computationally lightweight.

\begin{itemize}

\item \textbf{Phase 1 (exploration):} training proceeds from epoch 0 to epoch 10{,}000 using a cosine annealing scheduler with warm restarts (\emph{CosineAnnealingWarmRestarts}, $T_0=50$, $T_{\text{mult}}=2$, $\eta_{\min}=10^{-10}$). During this phase, additive white Gaussian noise (AWGN) is sampled uniformly from $\sigma \in [0,50]$.

\item \textbf{Phase 2 (refinement):} training continues from epoch 10{,}000 to epoch 20{,}000 starting from the best checkpoint obtained in Phase 1. 
A step scheduler (\texttt{StepLR}, \texttt{step\_size}=2000, $\gamma=0.6$) is used in this phase. 
The resulting models are referred to as \textit{flexible models}, as they are trained to operate across a wide range of noise levels.

\item \textbf{Phase 3 (noise-specific fine-tuning):} from each flexible model, specialized variants are obtained for fixed noise levels $\sigma \in \{5,15,25,35\}$ by further fine-tuning from epoch 20{,}000 up to epoch 30{,}000, using the same step scheduler as in Phase 2. 
These noise-specific models are referred to as \textit{fixed models}, as they are optimized for a particular noise level.

\end{itemize}

After each training phase, we select the checkpoint that achieves the highest average PSNR on the validation split. 
This model is then used to initialize the subsequent training stage, ensuring progressive refinement guided by validation performance.

We evaluate two parameterizations of DeepBM3D: a lightweight model with 890K parameters that surpasses classical BM3D, and a larger variant with 3.4M parameters that achieves further gains. 
For each parameterization, we obtain both \textit{flexible} models (trained across noise levels) and \textit{fixed} models (fine-tuned for specific noise levels), resulting in four configurations: lightweight–flexible, lightweight–fixed, heavy–flexible, and heavy–fixed. 
These variants form the basis for our experimental evaluation, which we compare against a range of existing methods described below.

\subsection{Baselines and Comparison Methods}

\noindent To assess the performance of DeepBM3D, we compare it against representative denoising methods spanning classical, hybrid, and deep learning paradigms. As a classical collaborative filtering reference, we include BM3D~\cite{dabov2007bm3d}, using the IPOL implementation~\cite{lebrun2012bm3d}, and a DCT-based denoiser (\texttt{DCTDenoiser}) employed as a baseline in DCT2Net~\cite{herbreteau2022dct2net}. 

To bridge the gap between classical priors and data-driven learning, we also consider DCT2Net and its variant DCT/DCT2net~\cite{herbreteau2022dct2net}, which extend classical DCT-based pipelines with trainable modules. These hybrid models retain part of the structure and interpretability of traditional methods while benefiting from optimization through learning. We evaluate them using the official pretrained weights released by the authors.

Finally, we include the deep learning method FFDNet~\cite{zhang2018ffdnet}, a widely used denoiser trained end-to-end for grayscale AWGN removal with a noise-level map as input. For consistency, we also rely on the official pretrained weights provided by the authors.

All baselines are evaluated under identical conditions using fixed noise levels $\sigma \in \{5, 15, 25, 35\}$. Whenever supported, models are provided with the exact noise-level map to ensure fair comparisons. This setup allows us to attribute performance differences to the models themselves rather than to discrepancies in evaluation protocol.


\section{Results and Discussion}
\label{sec:results}

\noindent In this section, we evaluate the performance of DeepBM3D through quantitative and visual comparisons against baseline methods. We further analyze the impact of model capacity, training strategy, and architectural design choices through ablation studies.

\subsection{Quantitative Results}

\noindent Average PSNR results across four benchmark datasets and four standard noise levels are reported in Table~\ref{tab:quant_results}. DeepBM3D consistently improves upon classical and hybrid baselines, including BM3D, BM3D-Net, and the DCT2Net family, across most evaluated settings. This confirms the benefit of replacing handcrafted or partially learned components with fully differentiable modules while preserving the structural principles of collaborative filtering pipelines.

Compared to BM3D-Net, which also follows a BM3D-inspired design, DeepBM3D achieves more stable performance across noise levels. While BM3D-Net performs competitively around its reference noise level ($\sigma=15$), its performance degrades significantly for noise levels outside this range. In contrast, DeepBM3D maintains consistent behavior across all noise levels, both in its flexible and noise-specific variants.

Compared to FFDNet, which is a powerful, fully end-to-end learned model, DeepBM3D achieves competitive performance, particularly at low-to-moderate noise levels. In several cases (e.g., $\sigma = 5$ and $\sigma = 15$), our heavy fixed variant matches or slightly exceeds FFDNet, demonstrating the strength of our modular design despite the structural constraints imposed by collaborative transform-domain filtering. Notably, all predictions in DeepBM3D are produced through element-wise filtering in the transform domain, in contrast to the unconstrained nature of FFDNet.

At higher noise levels ($\sigma \geq 25$), FFDNet maintains a more significant advantage, particularly on natural datasets such as Kodak and McMaster. This highlights the limitations of strongly structured filtering architectures under extreme noise conditions, where larger receptive fields and more expressive filtering become advantageous. Nevertheless, DeepBM3D remains competitive, and its interpretability and modularity make it an attractive alternative.

Fine-tuning for fixed noise levels leads to consistent gains across all variants, with improvements of up to 0.2~dB compared to flexible training. The heavy fixed model achieves the highest scores overall, surpassing FFDNet in some settings and narrowing the gap in others. The lightweight models also perform remarkably well, showing that most performance gains can be achieved with compact architectures that retain the efficiency and modularity of collaborative filtering designs.

\begin{table*}[!htbp]
    \centering
    
\caption{
Average PSNR results (dB) on multiple datasets and noise levels. 
The table compares our proposed DeepBM3D variants (light/heavy, fixed/flexible) against classical (BM3D, DCTDenoiser), hybrid (BM3D-Net, DCT-based), and deep learning baselines (FFDNet).
The best and second-best results for each dataset and noise level are highlighted in \best{bold} and \second{underlined}, respectively.
}
\label{tab:quant_results}
    
\begin{tabular}{lccccc}
\toprule
 & $\sigma$ & BSD100 & Kodak & McM & Set14 \\
\midrule
DCTDenoiser & 5 & 36.62 & 37.47 & 38.23 & 37.10 \\
DCT2Net &   & 37.11 & 38.07 & 38.76 & 37.52 \\
DCT/DCT2net &   & 37.06 & 37.93 & 38.73 & 37.49 \\
BM3D-Net\textsuperscript{a}     &   & -     & -     & -     & -     \\
BM3D &   & 37.29 & 38.19 & 39.09 & 37.89 \\
FFDNet &   & 37.49 & 38.42 & 39.39 & 37.98 \\ \addlinespace 
DeepBM3D light fixed &   & \second{37.69} & \second{38.65} & \second{39.67} & \second{38.38} \\
DeepBM3D light flexible &   & 37.34 & 38.35 & 39.33 & 37.59 \\
DeepBM3D heavy fixed &   & \best{37.72} & \best{38.69} & \best{39.71} & \best{38.42} \\
DeepBM3D heavy flexible &   & 37.43 & 38.44 & 39.43 & 37.69 \\ \midrule
DCTDenoiser & 15 & 29.73 & 31.17 & 32.12 & 31.00 \\
DCT2Net &   & 30.84 & 32.31 & 33.21 & 32.02 \\
DCT/DCT2net &   & 30.71 & 32.12 & 33.12 & 31.97 \\
BM3D-Net     &   & 31.08 & 32.48 & 33.57 & 32.30 \\
BM3D &   & 30.69 & 32.20 & 33.29 & 32.16 \\
FFDNet &   & \second{31.36} & \second{32.86} & \second{34.00} & 32.70 \\ \addlinespace 
DeepBM3D light fixed &   & 31.35 & 32.85 & 33.97 & \second{32.82} \\
DeepBM3D light flexible &   & 31.14 & 32.68 & 33.80 & 32.58 \\
DeepBM3D heavy fixed &   & \best{31.40} & \best{32.92} & \best{34.05} & \best{32.90} \\
DeepBM3D heavy flexible &   & 31.22 & 32.77 & 33.90 & 32.70 \\ \midrule
DCTDenoiser & 25 & 27.24 & 28.83 & 29.62 & 28.50 \\
DCT2Net &   & 28.40 & 29.96 & 30.78 & 29.65 \\
DCT/DCT2net &   & 28.26 & 29.80 & 30.68 & 29.61 \\
BM3D-Net     &   & 28.49 & 29.93 & 30.88 & 28.95 \\
BM3D &   & 28.20 & 29.87 & 30.83 & 29.79 \\
FFDNet &   & \best{28.94} & \best{30.57} & \best{31.62} & \second{30.44} \\ \addlinespace 
DeepBM3D light fixed &   & 28.85 & 30.46 & 31.48 & 30.44 \\
DeepBM3D light flexible &   & 28.70 & 30.34 & 31.35 & 30.28 \\
DeepBM3D heavy fixed &   & \second{28.90} & \second{30.52} & \second{31.54} & \best{30.53} \\
DeepBM3D heavy flexible &   & 28.77 & 30.42 & 31.45 & 30.39 \\ \midrule
DCTDenoiser & 35 & 25.86 & 27.49 & 28.10 & 26.97 \\
DCT2Net &   & 26.94 & 28.50 & 29.18 & 28.08 \\
DCT/DCT2net &   & 26.81 & 28.39 & 29.08 & 28.05 \\
BM3D-Net\textsuperscript{a}     &   & 24.32 & 22.87 & 27.01 & 23.71\\
BM3D &   & 26.73 & 28.43 & 29.27 & 28.26 \\
FFDNet &   & \best{27.50} & \best{29.17} & \best{30.09} & \second{28.96} \\ \addlinespace 
DeepBM3D light fixed &   & 27.37 & 29.01 & 29.88 & 28.88 \\
DeepBM3D light flexible &   & 27.24 & 28.89 & 29.76 & 28.74 \\
DeepBM3D heavy fixed &   & \second{27.41} & \second{29.07} & \second{29.95} & \best{28.98} \\
DeepBM3D heavy flexible &   & 27.30 & 28.97 & 29.87 & 28.87 \\
\bottomrule
\end{tabular}    

\begin{flushleft}
\footnotesize
\textsuperscript{a}BM3D-Net is pretrained at $\sigma=15$. Although it accepts different noise levels as input via rescaling, its performance degrades for noise levels far from this reference point (e.g., $\sigma\in\{5,35\}$).
\end{flushleft}

\end{table*}


\subsection{Qualitative Results}
\begin{figure*}
    \centering
    \includegraphics[width=\linewidth]{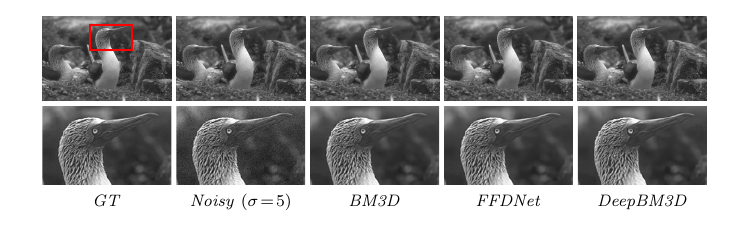}
    \caption{Qualitative comparison on \texttt{BSD100} at $\sigma\!=\!5$.
    Top: full images (GT annotated with zoom region). Bottom: identical zoomed region across methods.}
    \label{fig:qual_BSD100_5_004}
\end{figure*}

\begin{figure*}
    \centering
    \includegraphics[width=\linewidth]{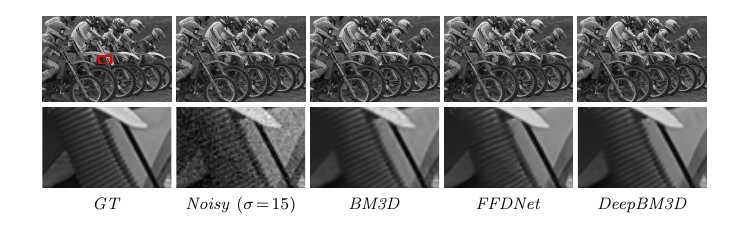}
    \caption{Qualitative comparison on \texttt{Kodak} at $\sigma\!=\!15$.
    Top: full images (GT annotated with zoom region). Bottom: identical zoomed region across methods.}
    \label{fig:qual_Kodak_15_005}
\end{figure*}

\begin{figure*}
    \centering
    \includegraphics[width=\linewidth]{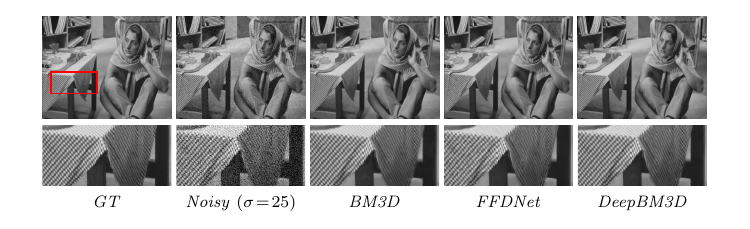}

    \caption{Qualitative comparison on \texttt{Set14} at $\sigma\!=\!25$.
    Top: full images (GT annotated with zoom region). Bottom: identical zoomed region across methods.}
    \label{fig:qual_Set14_25_002}
\end{figure*}

\begin{figure*}
    \centering
    \includegraphics[width=\linewidth]{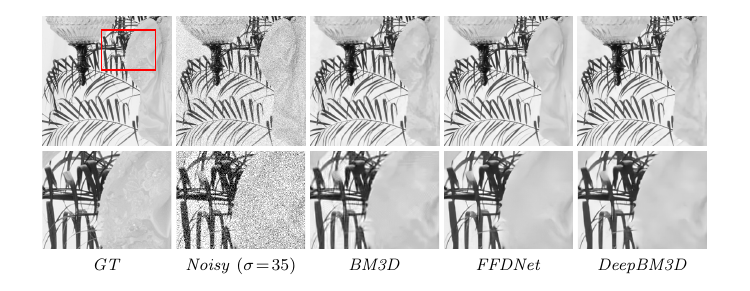}
    \caption{Qualitative comparison on \texttt{McMaster} at $\sigma\!=\!35$.
    Top: full images (GT annotated with zoom region). Bottom: identical zoomed region across methods.}
    \label{fig:qual_McMaster_35_004}
\end{figure*}

\noindent We complement the quantitative evaluation with qualitative comparisons at representative noise levels. Figures~\ref{fig:qual_BSD100_5_004}, \ref{fig:qual_Kodak_15_005}, \ref{fig:qual_Set14_25_002}, and \ref{fig:qual_McMaster_35_004} illustrate visual results for $\sigma\in\{5,15,25,35\}$, respectively.

Across the evaluated noise levels, DeepBM3D consistently outperforms BM3D, particularly at higher noise intensities where BM3D tends to produce structured artifacts such as blockiness or spurious lines. Our method preserves a cleaner appearance without introducing such distortions.

At low noise ($\sigma=5$, Figure~\ref{fig:qual_BSD100_5_004}), DeepBM3D achieves results on par with FFDNet in terms of both noise removal and detail preservation. At low-to-moderate noise levels ($\sigma=15$, Figure~\ref{fig:qual_Kodak_15_005}), our model shows clear advantages in reconstructing repetitive textures and high-frequency details. This is visible, for example, in the fine structures of the motorcycle spokes and forks.
Similar behavior on high-frequency repetitive details can be observed in Figure~\ref{fig:qual_Set14_25_002}.

At higher noise levels ($\sigma=35$, Figure~\ref{fig:qual_McMaster_35_004}), FFDNet retains a slight edge in preserving extremely fine details, yet DeepBM3D remains visually competitive, offering high-quality denoising with faithful structure reconstruction and effective suppression of noise. Overall, the combination of DCT-domain filtering and patch grouping yields visually compelling reconstructions across a wide range of noise conditions.

 \subsection{Texture Reconstruction Analysis}

 \noindent
 To further analyze the behavior of the proposed method, we evaluate its performance on structured textures from the Brodatz dataset. These images exhibit strong periodic patterns and non-local self-similarity, making them particularly suitable for assessing methods based on patch grouping and transform-domain filtering.

\begin{figure}
     \centering
     \includegraphics[width=\linewidth]{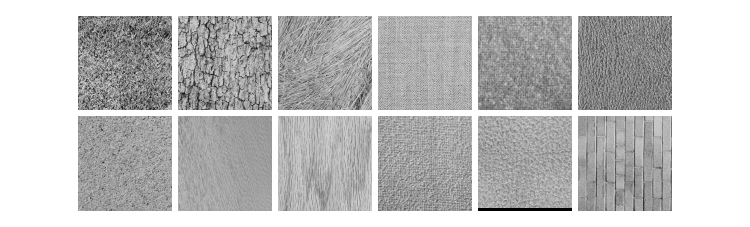}
     
     \caption{Brodatz textures used for building Table~\ref{tab:brodatz}.}
     \label{fig:brodatz}
 \end{figure}

 Quantitative results are reported in Table~\ref{tab:brodatz}.   Figure~\ref{fig:brodatz} illustrates the texture images used for building this table.  DeepBM3D consistently achieves superior PSNR values compared to FFDNet across all evaluated noise levels, which is not the case for general images as reported in Table ~\ref{tab:quant_results}.  The differences between the light and heavy versions are also narrower in the case of textures.  These results highlight the advantage of combining non-local grouping and transform-domain filtering within a learnable framework for images with strong  repetitive and oscillatory structure.

 \begin{table}
     \centering
     \caption{
     PSNR (dB) on Brodatz textures for different noise levels. 
     We compare classical (BM3D), deep learning (FFDNet), and the proposed DeepBM3D (\textit{light and heavy fixed}) model.
     }
     \label{tab:brodatz}
    
     \begin{tabular}{@{}lccc@{}}
         \toprule
          $\sigma$ & 15  & 25 & 35  \\
          \midrule
          BM3D &    29.00 & 26.52 & 25.05 \\
          FFDNet     &   29.32 &26.80 & 25.30 \\
          DeepBM3D Light &      \second{29.44} &  \second{26.91}  &\second{25.38}\\ 
         DeepBM3D Heavy &     {\bf 29.48} & {\bf 26.93} & {\bf 25.40} \\

          \bottomrule
     \end{tabular}
 \end{table}

\subsection{Ablation Studies}

We conduct ablation experiments to analyze the contribution of key design choices in \textit{DeepBM3D}.
These studies isolate the effect of individual components, such as the number of stages and the patch grouping configuration, providing insight into how each element influences denoising performance and visual quality.

\paragraph{Effect of the second stage}

\begin{figure}
    \centering
    \includegraphics[width=\linewidth]{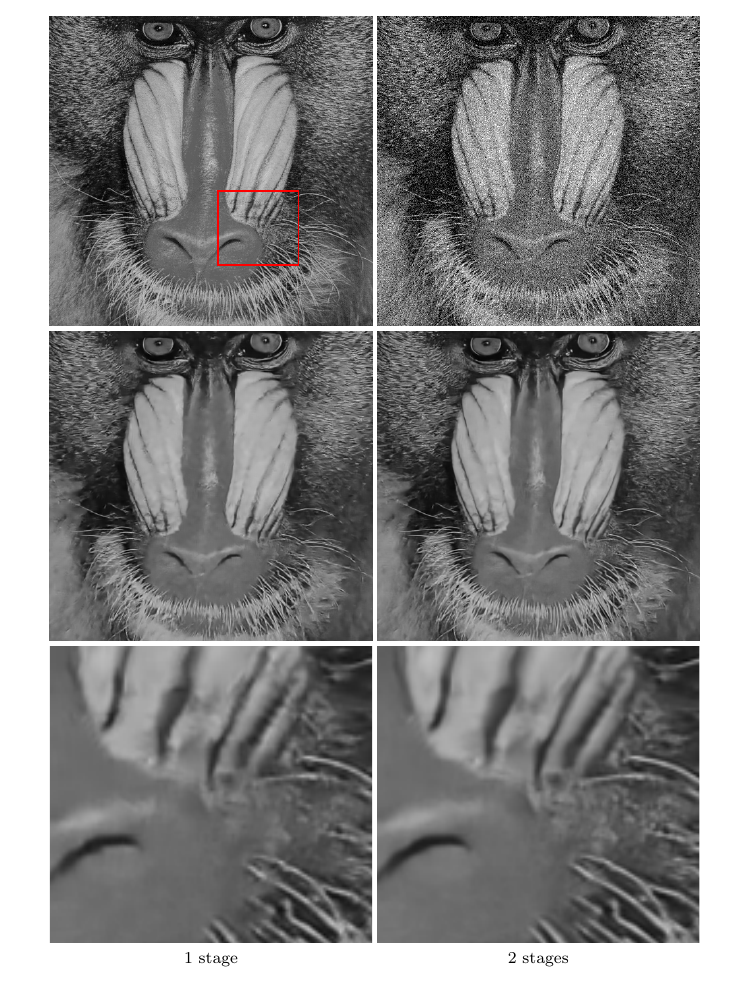}
    \caption{Comparison of the one- and two-stage variants of \textit{DeepBM3D} on \texttt{Set14} at $\sigma\!=\!35$. The two-stage model refines the first-stage output, yielding sharper and cleaner details in the zoomed region.}
    \label{fig:ablstage_Set14_35_001}
\end{figure}

To evaluate the contribution of the second stage in the DeepBM3D pipeline, we compare the performance of the single-stage and two-stage variants using the \textit{light flexible} configuration. Quantitative results are reported in Table~\ref{tab:stages}.
Across all datasets and noise levels, the two-stage model consistently outperforms the single-stage counterpart, confirming the benefit of iterative refinement.
The improvement becomes more noticeable at higher noise levels, where the second stage effectively attenuates residual noise and reduces structured artifacts that may appear after the first filtering.
Qualitative results are shown in Figure~\ref{fig:ablstage_Set14_35_001}. It can be seen that the second stage produces smoother and more coherent textures, mitigating aliasing effects while preserving the overall structure and natural appearance of the image.
These findings support the use of a two-stage refinement strategy within the proposed differentiable framework, as it enhances stability and visual consistency.

\begin{table}
    \centering
    \caption{
    Average PSNR (dB) on multiple datasets and noise levels, comparing single-stage and full two-stage pipelines for the DeepBM3D \textit{light flexible} variant. 
    }
    \label{tab:stages}
    \begin{tabular}{lccccc}
        \toprule
         & $\sigma$ & BSD100 & Kodak & McM & Set14 \\
         \midrule
         1 stage &  5  & 36.87 & 38.19 & 39.18 & 37.32 \\
         2 stages &    & 37.34 & 38.35 & 39.33 & 37.59 \\ \midrule
         1 stage &  15 & 30.86 & 32.45 & 33.58 & 32.37 \\
         2 stages &    & 31.14 & 32.68 & 33.80 & 32.58 \\ \midrule
         1 stage &  25 & 28.41 & 30.04 & 31.05 & 29.95 \\
         2 stages &    & 28.70 & 30.34 & 31.35 & 30.28 \\ \midrule
         1 stage &  35 & 26.96 & 28.55 & 29.41 & 28.33 \\
         2 stages &    & 27.24 & 28.89 & 29.76 & 28.74 \\
         \bottomrule
    \end{tabular}
\end{table}

\paragraph{Effect of candidate count ($N$)}

\begin{figure*}
    \centering
    \includegraphics[width=0.7\linewidth]{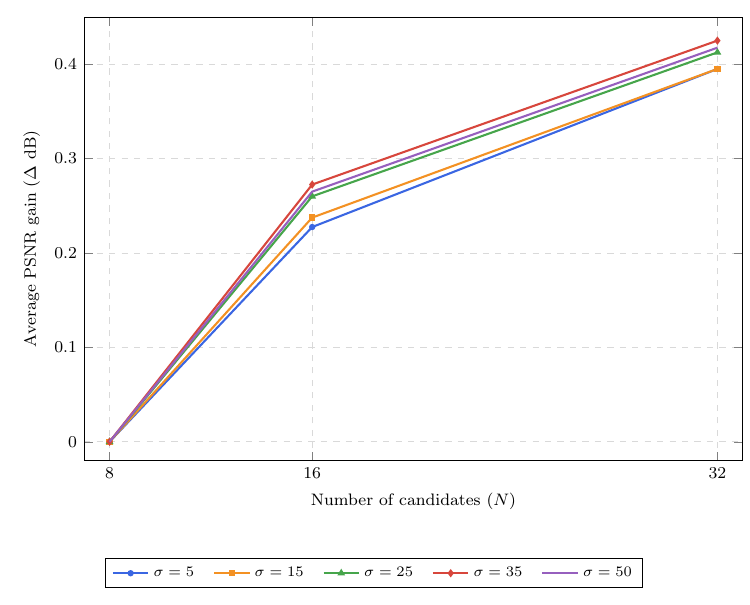}
        \caption{
    \textbf{Effect of the number of candidates.}
    Average PSNR gain ($\Delta$\,dB) achieved by varying the number of candidates ($N$)
    in the \textit{light flexible} DeepBM3D configuration, relative to the lightest setting with $N{=}8$, across different noise levels.
    }
    \label{fig:candidates}
\end{figure*}

The results in Table~\ref{tab:candidates}, together with the visual summary in Figure~\ref{fig:candidates}, show a clear and consistent improvement as the number of candidates increases. Across all datasets and noise levels, larger groups yield higher PSNR values, confirming that more extensive aggregation enhances the collaborative filtering stage. The effect is especially noticeable for low and medium noise levels ($\sigma{\leq}15$), where increasing $N$ from 8 to 32 provides gains of up to 0.5~dB, particularly on the McMaster and Kodak datasets. Although the relative improvement becomes smaller for higher noise levels, the trend remains monotonic across all cases.  

Among the evaluated datasets, McMaster exhibits the greatest sensitivity to the number of candidates, suggesting that richly textured and colorful images benefit the most from dense patch grouping. The corresponding plot in Figure~\ref{fig:candidates} illustrates this monotonic behavior, showing how each noise level consistently benefits from larger groups, although with diminishing returns. Moreover, the gains from $N{=}16$ to $N{=}32$ are not negligible, indicating that the model could achieve even higher performance with larger groups. However, increasing $N$ substantially raises the parameter count and computational cost during both training and inference.  

For this reason, we adopt $N{=}16$ as a balanced configuration for our experiments, offering an excellent trade-off between denoising quality and efficiency. Overall, these results highlight the flexibility of DeepBM3D to scale its accuracy depending on computational constraints and application needs.

\begin{table}
    \centering
    \caption{
    Average PSNR (dB) on multiple datasets and noise levels, comparing the effect of different candidate counts ($N=C{+}1$) in the DeepBM3D (\textit{light flexible}) variant. 
    }
    \label{tab:candidates}
    \begin{tabular}{lccccc}
        \toprule
         & $\sigma$ & BSD100 & Kodak & McM & Set14 \\
         \midrule
         $N=8$ &  5  & 37.06 & 38.17 & 39.10 & 37.37 \\
         $N=16$ &    & 37.34 & 38.35 & 39.33 & 37.59 \\
         $N=32$ &    & 37.52 & 38.49 & 39.51 & 37.76 \\ \midrule
         $N=8$ & 15  & 30.92 & 32.44 & 33.55 & 32.34 \\
         $N=16$ &    & 31.14 & 32.68 & 33.80 & 32.58 \\
         $N=32$ &    & 31.28 & 32.83 & 33.96 & 32.76 \\ \midrule
         $N=8$ & 25  & 28.47 & 30.09 & 31.09 & 29.98 \\
         $N=16$ &    & 28.70 & 30.34 & 31.35 & 30.28 \\
         $N=32$ &    & 28.83 & 30.48 & 31.51 & 30.46 \\ \midrule
         $N=8$ & 35  & 27.00 & 28.64 & 29.49 & 28.41 \\
         $N=16$ &    & 27.24 & 28.89 & 29.76 & 28.74 \\
         $N=32$ &    & 27.35 & 29.03 & 29.93 & 28.93 \\ 
         \bottomrule
         
    \end{tabular}
\end{table}

\subsection{Limitations}

\begin{figure*}
    \centering
    \includegraphics[width=\linewidth]{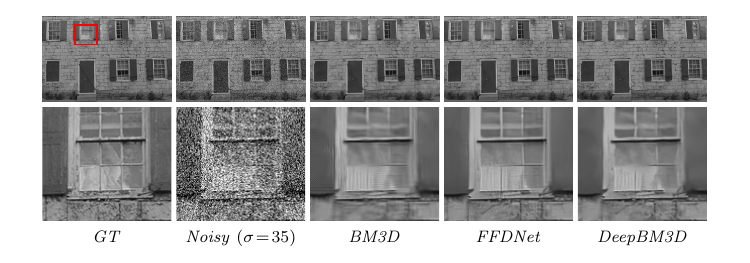}
        \caption{Qualitative comparison on \texttt{Kodak} at $\sigma\!=\!35$.
    Top: full images (GT annotated with zoom region). Bottom: identical zoomed region across methods.}
    \label{fig:qual_Kodak_35_001}
\end{figure*}

\noindent While DeepBM3D shows strong performance across most scenarios, its limitations become more apparent at higher noise levels. In such cases, the method tends to produce slightly blurrier results compared to fully end-to-end models such as FFDNet, as shown in Figure~\ref{fig:qual_Kodak_35_001}.

This behavior is expected, as our architecture inherits certain structural constraints from BM3D. First, it operates on relatively small 3D blocks constructed from $5\times5$ patches, limiting the spatial context available during filtering. Second, the grouping stage is restricted to 15 non-local candidates plus the reference patch, which may be insufficient under severe noise. Lastly, the core filtering operation consists of an element-wise multiplication between the original patches and learned weights, providing a more constrained representation compared to the deeper convolutional architectures used in modern denoisers. These factors collectively constrain the model’s capacity, especially under challenging high-noise conditions.

\section{Conclusion}
\label{sec:conclusion}

\noindent
In this work, we introduced DeepBM3D, a fully differentiable architecture for collaborative transform-domain image denoising inspired by classical BM3D-style pipelines. 
The proposed model combines non-local patch grouping, DCT-domain collaborative filtering, and multi-stage refinement within a unified trainable framework, replacing handcrafted operations with lightweight differentiable modules while preserving the structural principles of collaborative filtering methods.

Experimental results show that DeepBM3D consistently improves over classical and hybrid baselines and remains competitive with modern deep denoisers such as FFDNet, despite relying on significantly stronger structural constraints. 
In particular, the proposed architecture performs especially well on repetitive and oscillatory textures, suggesting that non-local grouping and DCT-domain filtering remain highly effective priors for structured image content.

At the same time, our experiments reveal the limitations of strongly structured transform-domain filtering architectures. 
Under high noise conditions, restricted patch groups and localized filtering become less expressive than unconstrained deep models with larger receptive fields and higher representational capacity. 
These observations provide insight into both the strengths and the limits of collaborative DCT-based denoising within modern learning frameworks.

Overall, our results suggest that a substantial part of image denoising performance can still be explained through classical principles such as transform sparsity and non-local self-similarity. 
Rather than replacing these ideas, deep learning can enhance them through differentiable optimization, enabling compact and interpretable architectures that remain competitive with significantly larger neural models. 
Future work will investigate how far these structured approaches can be pushed through richer grouping strategies, adaptive transforms, and more expressive collaborative filtering mechanisms, particularly in challenging high-noise and real-world restoration scenarios.

\backmatter

\bmhead{Funding}

This work was supported by MCIN/AEI/10.13039/501100011033 and FEDER, European Union, under grant PID2021-125711OB-I00, and by the Spanish Ministry of Universities under grant FPU24/02805.

The funders had no role in the study design; in the collection, analysis, or interpretation of data; in the writing of the manuscript; or in the decision to submit the article for publication.

\bmhead{Acknowledgment}
The authors gratefully acknowledge the computer resources at Artemisa and the technical support provided by the Instituto de Fisica Corpuscular, IFIC(CSIC-UV). Artemisa is co-funded by the European Union through the 2014-2020 ERDF Operative Programme of Comunitat Valenciana, project IDIFEDER/2018/048. Likewise, the authors acknowledge the computer resources at Talaia and the technical support provided by the Balearic Islands Center for Supercomputing and Artificial Intelligence (BSAI) at the Universitat de les Illes Balears (UIB).

\bmhead{Author contributions}
C.C. developed and implemented the proposed method, conducted the experiments, analyzed the results, prepared the figures and tables, and wrote the main manuscript text. J.N. and A.B. supervised the research, contributed to the conceptual and methodological development of the work, provided guidance throughout the study, and critically reviewed and edited the manuscript. A.B. acquired funding and provided project administration. All authors discussed the results, contributed to the final version of the manuscript, and approved the submitted version.

\bmhead{Data availability}

The datasets, trained models, and code generated and/or analyzed during the current study are available from the corresponding author upon reasonable request.

\bmhead{Competing interests}

The authors declare no competing interests.


\bibliography{biblio}

\end{document}